%% file: main.tex
\documentclass[letterpaper, 10 pt, conference]{ieeeconf}  

\IEEEoverridecommandlockouts                              

\usepackage[utf8]{inputenc} 
\usepackage[T1]{fontenc}    
\usepackage{hyperref}       
\usepackage{url}            
\usepackage{booktabs}       
\usepackage{amsfonts}       
\usepackage{cite}
\usepackage{color}

\usepackage{amsmath} 
\usepackage{amssymb}  
\usepackage{exscale}
\usepackage{graphicx}
\usepackage{balance}
\usepackage{graphicx}
\usepackage{lscape}
\usepackage{longtable}
\usepackage{caption}
\usepackage[table,xcdraw]{xcolor}
\usepackage{multicol}
\usepackage{multirow}
\usepackage{subcaption}

\title{\LARGE \bf
Social Robot  Navigation through Constrained Optimization: a Comparative Study of Uncertainty-based Objectives and Constraints
}

\author{Timur Akhtyamov$^{1}$, Aleksandr Kashirin$^{1}$, Aleksey Postnikov$^{1,2}$, Gonzalo Ferrer$^{1}$
\thanks{$^{1}$The authors are with Skolkovo Institute of Science and Technology (Skoltech), Center for AI Technology. Corresponding e-mail:
         {\tt\small timur.akhtyamov@skoltech.ru}}%
\thanks{$^{2}$The author is with Sber Robotics group \newline 979-8-3503-0704-7/23/\$31.00 ©2023 IEEE}%
}

\begin{document}

\maketitle

\begin{abstract}

This work is dedicated to the study of how uncertainty estimation of the human motion prediction can be embedded into constrained optimization techniques, such as Model Predictive Control (MPC) for the social robot navigation.
We propose several cost objectives and constraint functions obtained from the uncertainty of predicting pedestrian positions and related to the probability of the collision that can be applied to the MPC, and all the different variants are compared in challenging scenes with multiple agents.
The main question this paper tries to answer is: what are the most important uncertainty-based criteria for social MPC? For that, we evaluate the proposed approaches with several social navigation metrics in an extensive set of scenarios of different complexity in reproducible synthetic environments.
The main outcome of our study is a foundation for a practical guide on when and how to use uncertainty-aware approaches for social robot navigation in practice and what are the most effective criteria.

\end{abstract}

\input{intro}
\input{related}
\input{method}
\input{eval}
\input{conclusion}


\end{document}

%% file: intro.tex
\section{Introduction}\label{sec_intro}

Social robot navigation remains a difficult problem since navigating in a socially acceptable manner, in a dynamic and complex environment, is often unpredictable and uncertain mostly due to its human nature. This involves not only avoiding obstacles but also interacting with humans in a way that is natural, safe, and comfortable. 

One of the main challenges is that human behaviour is ambiguous and difficult to predict. People may move in unexpected ways, change direction suddenly, or give non-verbal cues that are difficult for robots to interpret. Fortunately, with modern techniques now it is possible to predict accurately and with a correct measure of the inherent uncertainty \cite{covariance-net}. In addition, social norms and conventions vary between cultures and contexts, making it difficult to develop a one-size-fits-all approach.


Finally, safety is a critical concern in social navigation, as robots must avoid collisions and other hazards while navigating in close proximity to humans. This requires advanced planning and control algorithms that can take into account the robot's own capabilities and limitations, as well as those of the people in the environment. It is unclear which are the dominant criteria in social robot navigation, and our initial hypothesis is that accurate uncertainty prediction should play a fundamental role on the social navigation task.

MPC is one of the world's industrial standards for the variety of control and planning tasks, especially in robotics. Modern MPC solutions are built on top of the efficient solvers that achieve real-time or near-real-time performance in various deterministic settings.
Constrained optimization techniques employed by MPC allow to leverage different navigation objectives and constrains, for instance, distance to goal, probability of collision or deterministic geometric collision constraints.



In this work, we propose to study how pedestrians trajectory prediction uncertainty can be embedded into MPC-based planning via various objectives and constraints derived from the uncertainty in the environment, and how it influences in practice the performance of the controller. The main contributions of the paper are:
\begin{itemize}
    \item Several uncertainty-unaware and uncertainty-aware MPC designs that incorporate CovarianceNet-based approach \cite{covariance-net} for pedestrian trajectory prediction;
    \item Extensive evaluation of the proposed approaches in simulation environments with practice-oriented conclusions;
    \item Introduction of the novel simulation environment targeted for social robot navigation tasks.
\end{itemize}





%% file: related.tex
\section{Related Works}\label{sec_related}

\subsection{Social robot navigation approaches}


Generally, social robot navigation problem has been studied for several decades, and variety of approaches have been proposed \cite{kruse2013human, charalampous2017recent, mavrogiannis2021core}. Methods based on the enhancement of the classical path planning \cite{kollmitz2015time, rios2011understanding, chi2017risk, majd2021safe} are built on top of the algorithms like A*, RRT or RRT*. Adaptivity to the pedestrian dynamics is achieved by using time-based variations of those algorithms, dynamic cost maps that are built using pedestrians motion prediction and socially-aware transition or steering functions.

Optimization-based methods employ advances in non-linear programming to generate a sequence of safe robot control inputs. These methods first of all include MPC schemes adapted to the social navigation and dynamic collision avoidance \cite{chen2021interactive, brito2021go, poddar2023crowd}. The core idea is to use an external pedestrian trajectory prediction method and embed its output into the cost function or constraints.

Some authors also relate reaction-based methods like Social Force Model \cite{helbing1995social, moussaid2010walking, HSFM, ferrer2019akp} and velocity obstacles \cite{orca} to the possible social navigation approaches. But in practice, those methods usually applied as supervisors for pre-training learning-based models or combined with optimization-based or learning-based approaches.

With the rising popularity of Deep Learning, learning-based social navigation, especially Reinforcement Learning (RL)-based methods have become their own direction in robotics \cite{chen2017socially, chen2019crowd, chen2020relational, liu2022socially}. For today, main directions in the RL-based social navigation research are modeling interaction between pedestrians and robot \cite{chen2019crowd, chen2020relational, liu2022socially}, efficient usage of the pedestrians motion prediction by RL policy \cite{chen2020relational, liu2022socially, sathyamoorthy2020densecavoid} and combination of the RL-based methods with non-learnable approaches \cite{brito2021go, patel2021dwa, xie2023drl}.

\subsection{Uncertainty-aware objectives and constraints} \label{sec_related_objectives}

The goal of incorporating into robot navigation in general is to minimize collision probability, directly or indirectly. A common approach is to model robot and pedestrian as circles (or spheres, if going to 3D), but calculating exact collision probability even for such simple representation is a challenging problem \cite{cooper2020toolbox, du2011probabilistic}.

One way of tackling this issue is the chance constraint which gives approximate bounds on collision up to fixed probability. Several groups of chance constraints are present in the literature. The first group is based on approximation of the collision probability or finding its upper bound \cite{du2011probabilistic, althoff2010probabilistic, park2017efficient, park2018fast, thomas2022probabilistic}. The second group represents dynamic obstacles as circles or ellipses whose sizes derived via Gaussian level-sets of some fixed probability \cite{majd2021safe, schwarting2017parallel, busch2022gaussian}.

Another way of incorporating uncertainty into planning is using the concept of risk introduced in \cite{majumdar2020should}. According to \cite{majumdar2020should}, risk is defined as a mapping of the cost random variable to a real number that should follow a set of axioms. Most popular risk metrics that can be found in the literature are Expected Cost \cite{chen2021interactive}, Conditional Value at Risk (CVaR) \cite{novin2021risk, cai2022probabilistic, triest2023learning} and Mean-Variance \cite{kahn2017uncertainty}.

Recent works also made steps towards uncertainty-awareness in RL via risk-aware RL \cite{tamar2015policy, zhang2021mean, jaimungal2022robust} and Distributional RL \cite{bellemare2017distributional, dabney2018distributional}, but application of those method for social robot navigation problem is not well-studied problem yet.

\subsection{Uncertainty-aware trajectory prediction}
Uncertainty-aware trajectory prediction has been an active research area in robotics and autonomous navigation, particularly for applications involving social interactions. Traditional approaches to trajectory prediction rely on deterministic models \cite{helbing1995social}, which may not account for the inherent uncertainty in the environment and the behavior of other agents. To address this issue, several recent works have proposed uncertainty-aware prediction models that explicitly model the uncertainty in the trajectory estimation \cite{gilles2022gohome, covariance-net, salzmann2020trajectron++}. In this work we use a variation of the CovarianceNet\cite{covariance-net} model as an explicit method for uncertainty prediction in pedestrian trajectory estimation. 
In sake of simplicity and computational efficiency, our implementation is not using the Conditional Variational Autoencoder (CVAE) part of the original model. While the CVAE has been shown to produce diverse and realistic trajectories, we expect it would suffice to achieve the desired performance to omit the CVAE from our implementation of CovarianceNet. Also, as an underlying trajectory prediction method for CovarianceNet, the Constant Velocity (CV) model is used. Despite its extreme simplicity, in practice CV often produces results comparable to more sophisticated models in both prediction and navigation tasks \cite{liu2022socially, scholler2020constant}.





%% file: method.tex
\section{Method}\label{sec_methods}

In this section, we present a comprehensive explanation of our proposed approaches that utilize MPC. For ease of reference, Table \ref{tab:variable-definition} is provided to define the main variables used throughout this section.


\begin{table}[]
\centering
\caption{Variable Definition Table.}
\label{tab:variable-definition}
\resizebox{\columnwidth}{!}{%
\begin{tabular}{lll}
\toprule
\multicolumn{1}{l}{\textbf{Variable}}  & \multicolumn{1}{l}{\textbf{Definition}}             \\
\midrule
\(N\)                                  &   number of pedestrians                             \\
$i=\{1, \ldots, N\}$                   &   pedestrian index                                  \\
$H$                                  &   number of receding horizon steps                  \\
$k=\{0,1, \ldots, H-1\}$             &   receding horizon step index                       \\
$\Delta t$                           &   receding horizon time step interval, $\left[s\right]$        \\
$T^{sim}$                              &   number of simulation steps                  \\
$\Delta t^{sim}$                       &   simulation time step interval, $\left[s\right]$             \\
$H^{ghost}$                          &   number of receding horizon steps                  \\
                                       &   to track ghost pedestrians \\
$r^{rob}$                              &   robot circumference radius, $\left[m\right]$                 \\
$V^{S}$                                &   volume of the sphere used for \\
                                       &   Mahalanobis constraints, $\left[m^3\right]$      \\
$r^{ped}$                              &   pedestrian circumference radius, $\left[m\right]$            \\
$d^{safe}$                             &   safe distance between robot and                   \\
                                       &   pedestrian circumferences, $\left[m\right]$                  \\
$\varepsilon$                          &   target reach threshold, $\left[m\right]$ \\           
$\ell^{vis}$                           &   robot vision range, $\left[m\right]$ \\
$\varphi^{vis}$                        &   robot angle of view, $\left[rad\right]$ \\   
$\delta$                             &   adaptive margin constraint value \\
$x$                                  &   position along $x$ axis, $\left[m\right]$                 \\
$y$                                  &   position along $y$ axis, $\left[m\right]$               \\
$\theta \in \left[ -\pi; \pi \right)$ &   angular position, $\left[rad\right]$                         \\
$v$                                  &   linear velocity, $\left[\frac{m}{s}\right]$                  \\
$\omega$                             &   angular velocity, $\left[\frac{m}{s}\right]$                 \\
$\mathbf{r}_k = [x^{rob}_k, y^{rob}_k]^\top $                                               & robot position vector at step $k$\\
$\mathbf{r}_{target}$                                                                       & robot target position\\
$\mathbf{x}_k = [x^{rob}_k, y^{rob}_k, \theta^{rob}_k]^\top$                                & robot state vector at step $k$\\
$\mathbf{u}_k = [v^{rob}_k, \omega^{rob}_k]^\top$                                           & robot control vector at step $k$\\
$\bar{\mathbf{u}}_k = [v^{rob}_k, \omega^{rob}_k, \delta_{k}]^\top$                         & augmented robot control vector at step $k$\\
$\mathbf{p}_{k, i} = [x^{ped}_{k, i}, y^{ped}_{k, i}]^\top $                                & $i$-th pedestrian position vector at step $k$\\
$\Sigma_{k , i}$                                                                            & covariance of $i$-th pedestrian at $k$ step\\
$\lambda_{k , i}^{(1)}, \lambda_{k , i}^{(2)}$                                              & eigenvalues of the $\Sigma_{k , i}$\\
\(\gamma\)                                                                                    & number of standard deviations\\
\(a_{k , i}, b_{k , i}\)                                                                      & length of the ellipsoid constraint semi-axes\\
\(\psi_{k, i}\)                                                                               & rotation angle of the ellipsoid constraint\\
\(\text{Rot}(\psi)\)                                                                          & rotation matrix\\
\(P^{col}\)                                                                                   & collision probability threshold \\
\(Q_{\mathbf{u}}\)                                                                            & control input weight matrix \\
\(Q_{\bar{\mathbf{u}}}\)                                                                      & augmented control input weight matrix \\
\(Q_{\mathbf{r}}\)                                                                            & position unattainability factor \\
\(Q_{\mathbf{ED}}\)                                                                           & Euclidean distance cost weight \\
\(Q_{\mathbf{MD}}\)                                                                           & Mahalanobis distance cost weight \\
\(d^{\mathbf{ED}}_{k , i}\)                                                                   & Euclidean distance \\
                                                                                              & for $i$-th pedestrian at $k$ step \\
\(d^{\mathbf{MD}}_{k , i}\)                                                                   & Mahalanobis distance \\
                                                                                              & for $i$-th pedestrian at $k$ step \\
$\mathbb{W}$                                                                                  & position space \\
$\mathbb{X}$                                                                                  & state space \\
$\mathbb{U}$                                                                                  & action space \\
$\bar{\mathbb{U}}$                                                                            & augmented action space \\
\bottomrule
\end{tabular}         
}
\end{table}

\subsection{Robot Model}

 We first introduce the target robot model and system dynamics. The deterministic Markov decision process serves as a critical constraint that governs the behavior of the system. For this study, we selected the kinematic unicycle model of the robot, which can be represented as a discrete system:
\begin{equation} \label{eq:robot-dynamics-common}
     \begin{array}{@{}l}
          x^{rob}_{k+1} = x^{rob}_{k} + v^{rob}_k \cdot cos(\theta^{rob}_{k}) \cdot \Delta t, \\
          y^{rob}_{k+1} = y^{rob}_{k} + v^{rob}_k \cdot sin(\theta^{rob}_{k}) \cdot \Delta t, \\
          \theta^{rob}_{k+1} = \theta^{rob}_{k} + w^{rob}_k \cdot \Delta t.
     \end{array}
\end{equation}
It should be noted that in all proposed methods, the robot (ego-agent) and the pedestrians (agents) are modeled as circles with respective radii ($r^{rob}$ and $r^{ped}$). The control input vector consists of linear and angular velocities, denoted as \(\mathbf{u}_k = [v^{rob}_k, \omega^{rob}_k]^\top\).

\subsection{Model Predictive Control}

This section outlines the methods we have studied, all of which are based on MPC. MPC is an advanced control strategy that predicts the future behavior of a system using a mathematical model and a cost function as an optimization objective that encapsulates the target behavior of the agent. The cost function in our proposed methods consists of two parts: the stage cost (\ref{eq:stage-cost}) and the terminal cost (\ref{eq:terminal-cost}). The stage cost is accumulated at each stage of the prediction horizon up to the terminal step and includes the control input cost (\ref{eq:control-cost}) and the normalized target distance cost (\ref{eq:position-cost}), which was inspired by the cost function presented in \cite{brito2021go}.


The control input cost \ref{eq:control-cost} penalizes the usage of the control signal, which consists of the linear and angular velocities:
\begin{equation} \label{eq:control-cost}
    J_{k}^{\mathbf{u}}\left(\mathbf{u}_{k}\right) = \mathbf{u}^\top_{k} Q_{\mathbf{u}} \mathbf{u}_{k}.
\end{equation}

The normalized target distance cost (\ref{eq:position-cost}) penalizes the robot's deviation from the target position during the optimization process. This cost decreases as the robot gets closer to the target position at each iteration relative to its initial position at the beginning of the horizon:


\begin{equation} \label{eq:position-cost}
    J_{k}^{\mathbf{r}}\left(\mathbf{r}_{k}\right) = Q_{\mathbf{r}}
    \left(\frac{ \left\| \mathbf{r}_{k} - \mathbf{r}_{target} \right\|_2 } 
         { \left\| \mathbf{r}_0 - \mathbf{r}_{target} \right\|_2} \right)^2.
\end{equation}

The combination of the control input cost and the normalized target point distance cost results in the definition of the basic stage cost (\ref{eq:stage-cost}):

\begin{equation} \label{eq:stage-cost}
    J_{k} (\mathbf{u}_{k}, \mathbf{r}_{k}) = J_{k}^{\mathbf{u}}\left(\mathbf{u}_{k}\right) + J_{k}^{\mathbf{r}}\left(\mathbf{r}_{k}\right).
\end{equation}

In the terminal step of the optimization problem, we only penalize the robot's inability to reach the target position (\ref{eq:terminal-cost}):

\begin{equation} \label{eq:terminal-cost}
    J_{H} := J_k^{\mathbf{r}} (\mathbf{r}_k) \mid k = H.
\end{equation}

Combining all of the previously mentioned terms results in a basic MPC optimization problem, which we refer to as \textit{MPC} (\ref{eq:MPC}) in the Table \ref{tab:all-methods}:

\begin{equation} \label{eq:MPC}
    \begin{aligned}
        & \underset{\mathbf{r}_{1: H}, \mathbf{u}_{0: H-1}}{\text{min}}
        & &  \sum_{k=0}^{H-1} J_{k} (\mathbf{u}_{k}, \mathbf{r}_{k}) + J_{H}\left(\mathbf{r}_{H}\right) \\
        & \text{subject to}
        & & \mathbf{r}_{0}=\mathbf{r}(0) \\
        &&& \mathbf{u}_{k} \in \mathbb{U} \\
        &&& \mathbf{r}_{k} \in \mathbb{W}.
    \end{aligned}
\end{equation}

Currently, we have defined a basic MPC optimization problem that is suitable for navigation tasks. However, it does not consider pedestrians in the environment. In the following section, we discuss how we can incorporate pedestrians into the optimization problem, both considering and not considering uncertainty.

\subsubsection{Uncertainty-unaware}

We introduce a classical uncertainty-unaware approach commonly used in motion planning to account for obstacles, other agents, and environmental borders - the Euclidean distance. The Euclidean distance (ED) is a measure of the straight-line length between two points in Euclidean space \ref{eq:euclidean}: 


\begin{equation} \label{eq:euclidean}
    d^{\mathbf{ED}}_{k, i}\left(\mathbf{r}_{k}, \mathbf{p}_{k, i}\right)= \left\| \mathbf{r}_{k} - \mathbf{p}_{k, i} \right\|_2.
\end{equation}

It is often used in optimization problems to prevent controllers from colliding with obstacles by imposing a constraint on the distance between the ego-agent and other agents \cite{brito2021go}. However, in our approach, we also study the utilization of the Euclidean distance as a component \ref{eq:ED-cost} of the stage-cost function. This is usually referred to as penalty-based optimization:

\begin{equation} \label{eq:ED-cost}
    J_{k}^{\mathbf{ED}} (\mathbf{r}_{k}, \mathbf{p}_{k, 1:N}) =
    Q_{\mathbf{ED}} \sum_{i=0}^{N} \frac{1}{
             d^{\mathbf{ED}}_{k, i}\left(\mathbf{r}_{k}, \mathbf{p}_{k, i}\right)^2
    }.
\end{equation}

We refer to the optimization problem that includes Euclidean distance as an additional component of the stage-cost function as ED-MPC \ref{eq:ED-MPC}:

\begin{equation} \label{eq:ED-MPC}
        \begin{aligned}
            & \underset{\mathbf{r}_{1: H}, \mathbf{u}_{0: H-1}}{\text{min}}
            & &\sum_{k=0}^{H-1} \left( J_{k} (\mathbf{u}_{k}, \mathbf{r}_{k}) +J_{k}^{\mathbf{ED}} (\mathbf{r}_{k}, \mathbf{p}_{k, 1:N}) \right)
                + \\
            & && \phantom{aaaaaaaaaa} + J_{H}\left(\mathbf{r}_{H}\right) \\    
            & \text{subject to}
            & & \mathbf{r}_{0}=\mathbf{r}(0)  \\
            &&& \mathbf{u}_{k} \in \mathbb{U} \\
            &&& \mathbf{r}_{k} \in \mathbb{W}.
        \end{aligned}
\end{equation}

In order to use Euclidean distance as a constraint, an inequality must be introduced \ref{eq:EDC-inequality} to ensure that the safe distance between the ego-agent and pedestrians is not violated:

\begin{equation} \label{eq:EDC-inequality}
    d^{\mathbf{ED}}_{k, i}\left(\mathbf{r}_{k}, \mathbf{p}_{k, i}\right)^2 \geq (r^{rob}+r^{ped}+d^{safe})^2. \\
\end{equation}

An optimization problem that includes Euclidean distance as an inequality constraint is referred to as MPC-EDC (\ref{eq:MPC-EDC}):

\begin{equation} \label{eq:MPC-EDC}
    \begin{aligned}
        & \underset{\mathbf{r}_{1: H}, \mathbf{u}_{0: H-1}}{\text{min}}
        & &\sum_{k=0}^{H-1} J_{k} (\mathbf{u}_{k}, \mathbf{r}_{k})
            + J_{H}\left(\mathbf{r}_{H}\right) \\
        & \text{subject to}
        & & \mathbf{r}_{0}=\mathbf{r}(0)  \\
        &&& d^{\mathbf{ED}}_{k, i}\left(\mathbf{r}_{k}, \mathbf{p}_{k, i}\right)^2 \geq (r^{rob}+r^{ped}+d^{safe})^2 \\
        &&& \mathbf{u}_{k} \in \mathbb{U} \\
        &&& \mathbf{r}_{k} \in \mathbb{W}.
    \end{aligned}
\end{equation}

\subsubsection{Uncertainty-aware}

For the uncertainty-awareness, we first introduce approaches based on the Mahalanobis distance, which measures the distance between a point, e.g. robot position $\mathbf{r}_k$, and a distribution, e.g. predicted pedestrian position modeled as a Gaussian distribution with mean $\mathbf{p}_{k,i}$ and covariance matrix $\Sigma_{k,i}$ (which may include off-diagonal elements):
\begin{multline} \label{eq:mahalanobis}
         d^{\mathbf{MD}}_{k, i}\left(\mathbf{r}_{k}, \mathbf{p}_{k, i}, \Sigma_{k , i}\right) = \\=  \sqrt{\left( \mathbf{r}_{k} - \mathbf{p}_{k, i} \right)^\top \Sigma^{-1}_{k , i} \left( \mathbf{r}_{k} - \mathbf{p}_{k, i} \right)}.
\end{multline}
We propose to employ Mahalanobis distance as and alternative to the Euclidean distance that will allow MPC to capture uncertainty of the pedestrian trajectories prediction.



First, we propose to add the Mahalanobis distance as an additional component to the stage cost function. To do this, we compute a weighted sum of the inverse Mahalanobis distance to each pedestrian at each horizon step:
\begin{equation} \label{eq:MD-cost}
    J_{k}^{\mathbf{MD}} (\mathbf{r}_{k}, \mathbf{p}_{k, 1:N}) =
    Q_{\mathbf{MD}} \sum_{i=0}^{N} \frac{1}{
            d^{\mathbf{MD}}_{k, i}\left(\mathbf{r}_{k}, \mathbf{p}_{k, i}, \Sigma_{k , i}\right)^2
    }.
\end{equation}
This allows us to take into account for the uncertainty associated with each pedestrian's trajectory and adjust the cost function accordingly. An MPC controller that incorporates the Mahalanobis distance as an additional component to the stage cost function is referred to as MD-MPC.

The Mahalanobis distance can also be added as an inequality constraint to the optimization problem. Work \cite{du2011probabilistic} derives approximation for the collision probability for the spherical robot, and corresponding constraint expression for holding collision probability lower than given threshold probability $P^{col}$. We adopt this approximation to out problem where pedestrian position is uncertain and introduce following constraint:
\begin{equation} \label{eq:MDC-inequality}
    d^{\mathbf{MD}}_{k, i}\left(\mathbf{r}_{k}, \mathbf{p}_{k, i}, \Sigma_{k, i}\right)^2 \geq 2 \ln \left(\sqrt{\operatorname{det}\left(2 \pi \Sigma_{k , i}\right)} \frac{P^{col}}{V^{S}}\right), \\
\end{equation}
where $V^{S}$ is the volume of the sphere with radius $r^{rob} + r^{ped} + d^{safe}$, $P^{col}$ is the fixed collision probability threshold.

Along with Mahalanobis distance-based constraints, we propose another type of chance constraints, based on the idea of Gaussian iso-contours, proposed in \cite{schwarting2017parallel, busch2022gaussian}. Assuming that $\Sigma_{k,i}$ is the covariance of the $i$-th pedestrian's position at horizon step $k$ (which may include off-diagonal correlation terms), the parameters of the ellipsoid corresponding to the $\gamma$ standard deviations are derived. We calculate eigenvalues of the covariance matrix $\lambda_{k,i}^{(1)}$ and $\lambda_{k,i}^{(2)}$ which define ellipsoid semi-axes lengths and angle $\psi_{k,i}$ which define the rotation of the coordinate system related to the ellipsoid. Taking into account robot and pedestrian radii along with safe distance, length of the semi-axes of the bounding ellipsoid $a_{i,k}$ and $b_{i,k}$ are defined as:
\begin{equation} \label{eq:ELC-axes}
    \begin{bmatrix}
           a_{k,i} \\
           b_{k,i}
         \end{bmatrix} = \gamma \begin{bmatrix}
           \phantom{a} \sqrt{\lambda_{k,i}^{(1)}} \phantom{a} \\
           \phantom{a} \sqrt{\lambda_{k,i}^{(2)}} \phantom{a}
         \end{bmatrix} + r^{rob} + r^{ped} + d^{safe}.
\end{equation}
Final equation for the ellipsoid constraints has form:
\begin{multline} \label{eq:ELC-eq}
   \left( \mathbf{r}_k - \mathbf{p}_{k,i} \right)^\top \text{Rot}(\psi_{k, i})^\top \begin{bmatrix}
           \frac{1}{a_{k,i}^2} & 0 \\
           0 & \frac{1}{b_{k,i}^2}
         \end{bmatrix} \times \\ \times \text{Rot}(\psi_{k, i}) \left( \mathbf{r}_k - \mathbf{p}_{k,i} \right)  > 1,
\end{multline}
where $\text{Rot}(\psi_{k,i})$ defines the rotation matrix. This constraint holds that robot will not move inside the ellipsoid around pedestrian, and size of this ellipsoid is based on selected number of Gaussian standard deviations, and thus connected with collision probability.

We refer to a variant of the MPC that uses ellipsoid constraints as MPC-ELC. Stage cost and terminal cost are defined by equations \ref{eq:stage-cost} and \ref{eq:terminal-cost} correspondingly.

\subsubsection{Adaptive constraint}

We present an additional approach, called the \textit{adaptive constraint}, initially introduced in \cite{busch2022gaussian} via slack variable. The adaptive constraint approach involves introducing a new optimization variable, denoted as $\delta$, which is added to the augmented control input vector $\bar{\mathbf{u}}$. To properly formalize this approach, we introduce an augmented control input cost function, which replaces the original control input cost function (\ref{eq:control-cost}) in the optimization problem (\ref{eq:MPC}):
\begin{equation} \label{eq:augmented-control-cost}
    J_{k}^{\bar{\mathbf{u}}}\left(\bar{\mathbf{u}}_{k}\right) = \bar{\mathbf{u}}^\top_{k} Q_{\bar{\mathbf{u}}} \bar{\mathbf{u}}_{k}.
\end{equation}
Note that the initial robot model is not changed, and slack variable is added to the control vector for the ease of regularization. 

The adaptive constraint can be in conjunction with the Euclidean distance constraint to increase the safe distance between the ego-agent and the pedestrian:
\begin{equation} \label{eq:EDC-delta-inequality}
    d^{\mathbf{ED}}_{k, i}\left(\mathbf{r}_{k}, \mathbf{p}_{k, i}\right)^2 \geq (r^{rob}+r^{ped}+d^{safe})^2+\delta. \\
\end{equation}
The controller that incorporates both the adaptive constraint and the Euclidean distance constraint is referred to as MPC-AEDC. This approach provides additional flexibility in adjusting the safe distance between the ego-agent and the pedestrians, making it particularly useful in dynamic and uncertain environments. 

The adaptive constraint can also be used in conjunction with the Mahalanobis distance constraint to adjust the converted collision probability: 
\begin{equation} \label{eq:MDC-delta-inequality}
    d^{\mathbf{MD}}_{k, i}\left(\mathbf{r}_{k}, \mathbf{p}_{k, i}\right)^2 \geq 2 \ln \left(\sqrt{\operatorname{det}\left(2 \pi \Sigma_{k , i}\right)} \frac{P^{col}}{V^{S}}\right) + \delta. \\
\end{equation}
The controller that incorporates both the adaptive constraint and the Mahalanobis distance constraint is referred to as MPC-AMDC. This approach provides additional flexibility in adjusting the safety margin and collision probability.

We propose to apply adaptive constraint with ellipsoid constraints in a similar way to the original work \cite{busch2022gaussian}. We adjust the semi-axes of the bounding ellipsoid:
\begin{equation} \label{eq:ELC-axes-delta}
    \begin{bmatrix}
           a_{k,i} \\
           b_{k,i}
         \end{bmatrix} = \gamma \begin{bmatrix}
           \phantom{a} \sqrt{\lambda_{k,i}^{(1)}} \phantom{a} \\
            \phantom{a} \sqrt{\lambda_{k,i}^{(2)}} \phantom{a}
         \end{bmatrix} (1 - \delta) + r^{rob} + r^{ped} + d^{safe}.
\end{equation}

We refer to such a controller as MPC-AELC.

Table \ref{tab:all-methods} summarizes all the methods proposed in this paper and provides an overview of the design of each method.

\begin{table}[]
\centering
\caption{Summary of Methods.}
\label{tab:all-methods}
\resizebox{\columnwidth}{!}{%
\begin{tabular}{llll}
\toprule
\textbf{Controller Name} & \textbf{Cost Component} & \textbf{Constraint Type} \\
\midrule
ED-MPC                     & Euclidean (\ref{eq:ED-cost})                  & -           \\
ED-MPC-EDC                 & Euclidean (\ref{eq:ED-cost})                  & Euclidean (\ref{eq:EDC-inequality}) \\
ED-MPC-MDC                 & Euclidean (\ref{eq:ED-cost})                   & Mahalanobis (\ref{eq:MDC-inequality}) \\
MD-MPC-MDC                 & Mahalanobis (\ref{eq:MD-cost})                 & Mahalanobis (\ref{eq:MDC-inequality}) \\
MD-MPC-EDC                 & Mahalanobis (\ref{eq:MD-cost})                 & Euclidean (\ref{eq:EDC-inequality}) \\
ED-MPC-AEDC                & Euclidean (\ref{eq:ED-cost})                  & Adaptive Euclidean (\ref{eq:EDC-delta-inequality}) \\
MD-MPC-AEDC                & Mahalanobis (\ref{eq:MD-cost})                 & Adaptive Euclidean (\ref{eq:EDC-delta-inequality}) \\
MPC-AEDC                   & -                                               & Adaptive Euclidean (\ref{eq:EDC-delta-inequality})  \\
MPC-AMDC                   & -                                               & Adaptive Mahalanobis (\ref{eq:MDC-delta-inequality})\\
MPC-ELC-2                  & -                                              & Ellipsoid, $\gamma=2$ (\ref{eq:ELC-axes}, \ref{eq:ELC-eq})\\
MPC-ELC-3                  & -                                              & Ellipsoid, $\gamma=3$ (\ref{eq:ELC-axes}, \ref{eq:ELC-eq})\\
MPC-AELC-2                 & -                                              & Adaptive Ellipsoid, $\gamma=2$ (\ref{eq:ELC-axes-delta}, \ref{eq:ELC-eq})\\
MPC-AELC-3                 & -                                               & Adaptive Ellipsoid, $\gamma=3$ (\ref{eq:ELC-axes-delta}, \ref{eq:ELC-eq})\\
\bottomrule
\end{tabular}%
}
\end{table}

%% file: eval.tex
\section{Evaluation}
In this section, we provide a detailed description of our experimental setup, present and discuss the results of our experiments.

\subsection{Software and Datasets}
We utilized the \textit{do-mpc} framework \cite{do_mpc} for implementation\footnote{\url{https-//github.com/TimeEscaper/social_nav_baselines}} of all of the MPC-based controllers, which is built upon the \textit{CasADi} software package \cite{casadi} for nonlinear optimization and algorithmic differentiation. \textit{MUltifrontal Massively Parallel sparse direct Solver (MUMPS)} is used as a base solver for MPC problem. CovarianceNet implementation uses \textit{PyTorch} framework. As a simulation tool we have developed an open-source lightweight and flexible framework called \textit{PyMiniSim}\footnote{\url{https-//github.com/TimeEscaper/pyminisim}} (Fig. \ref{pyminisim}). Our implemented CovarianceNet model\footnote{\url{https-//github.com/alexpostnikov/CovarianceNet}} was trained on the subset of the Stanford Drone Dataset (SDD) \cite{robicquet2016learning}.

\subsection{Experimental Setup} \label{exp-setup}
We utilized the Headed Social Force Model (HSFM) \cite{HSFM}, an extension of a highly-regarded Social Force Model (SFM) \cite{SFM}, as a model for simulation of pedestrians behavior.

\begin{figure*}
  \centering
  \begin{subfigure}{0.25\textwidth}
    \includegraphics[width=\linewidth]{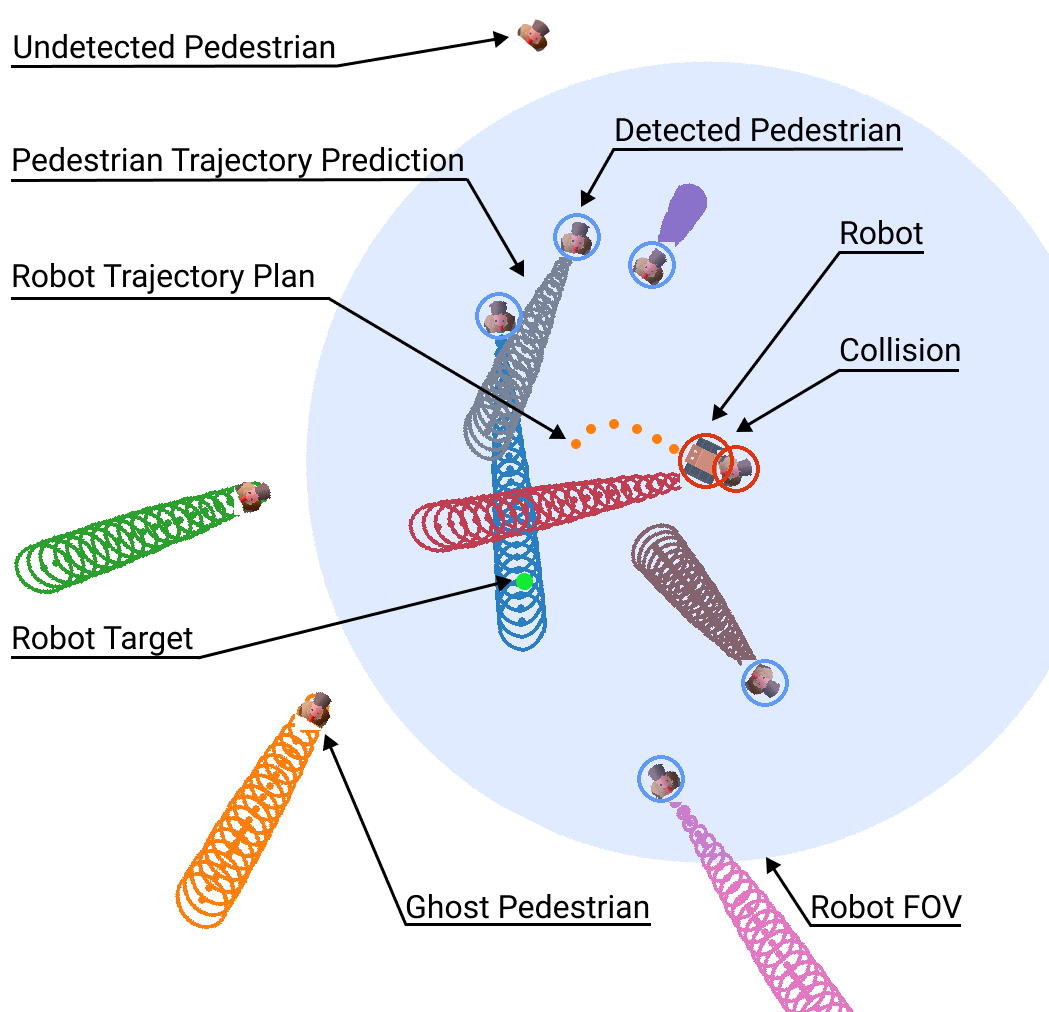}
    \caption{PyMiniSim environment.}
    \label{pyminisim}
  \end{subfigure}
  \begin{subfigure}{0.7\textwidth}
    \includegraphics[width=\linewidth]{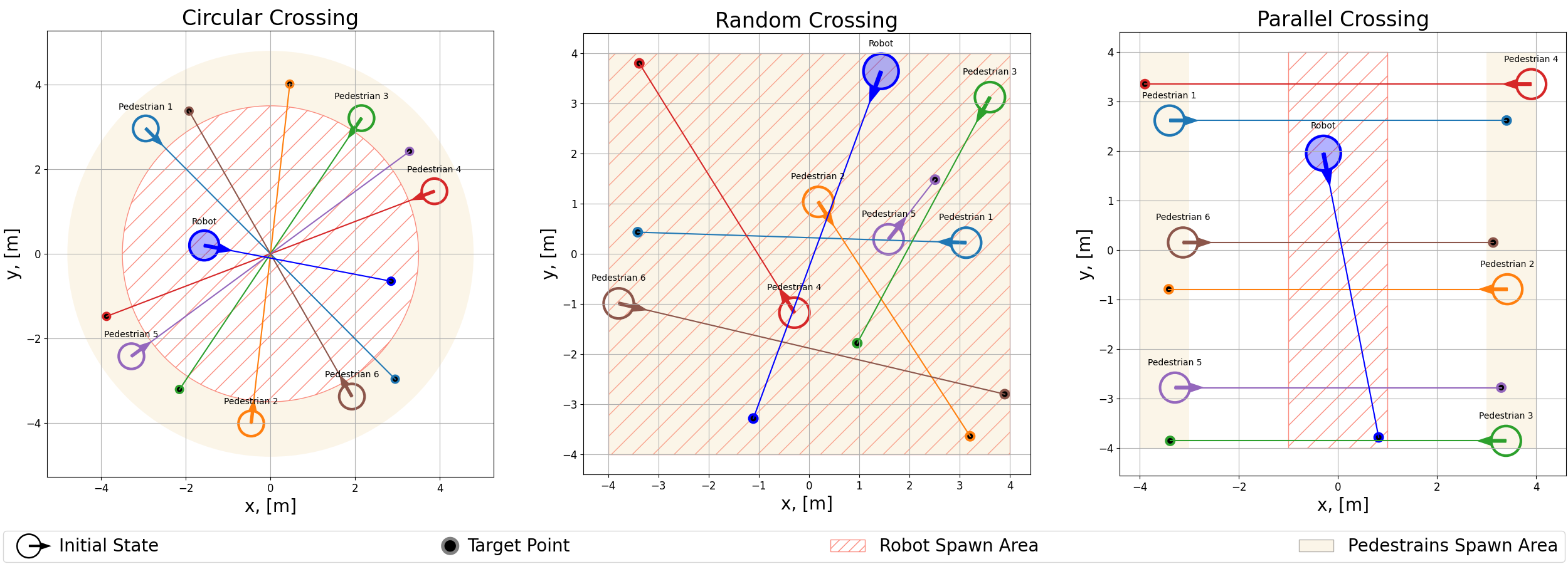}
    \caption{Scenario generation.}
    \label{scenario_generation}
  \end{subfigure}
  \caption{Experimental setup. In this study we used simulation environment based on PyMiniSim (Fig. \ref{pyminisim}) framework. We used scenarios represented in (Fig. \ref{scenario_generation}).}
  \label{fig:twopics}
\end{figure*}

To evaluate the effectiveness of the proposed methods, we designed and simulated three types of scenarios- \textit{circular crossing}, \textit{random crossing}, and \textit{parallel crossing}, as illustrated in Figure \ref{scenario_generation}. These scenarios were inspired by the work \cite{wang2022metrics}. For each scenario, we consider a set of scenes - configurations of number of pedestrians, initial pedestrians' poses, pedestrians' goal, initial robot pose and robot goal. Once pedestrians reach their goal positions, they oscillate between their initial and goal positions, resulting in continuous movement without stopping within the scene. Possible number of pedestrians varies from 3 to 8. For each number of the pedestrians, 100 scenes were generated, resulting in 600 scenes per scenario and 1800 scenes in total. Each of the controllers were evaluated on this set of scenes.

For evaluating the performance of the controllers, we utilized standard metrics such as \textit{Simulation steps to Target, $[\#]$}, which represents the time taken for the controller to reach the target position, \textit{Number of Collisions, $[\#]$}, and \textit{Number of Timeouts, $[\#]$}, which depict the cautiousness of the controller. Target is assumed to be reached by the robot if the following criterion holds:
\begin{equation} \label{eq-target-reach-criterion}
    \| \mathbf{r}_k - \mathbf{r}_{target} \|_2 - r^{rob} < \varepsilon.
\end{equation}
Here are the parameters that we used for the evaluation of the controllers:
\newcommand{\controlmatrix}{
    \left(
      \begin{smallmatrix}
        1 & 0 \\
        0 & 1 
      \end{smallmatrix}
    \right)
}
\newcommand{\augcontrolmatrix}{
    \left(
      \begin{matrix}
            0.005 & 0 & 1 \\
            0 & 0.005 & 0 \\
            0 & 0 & 100000
      \end{matrix}
    \right)
}
\begin{multicols}{2}
    \begin{itemize}
      \item[] $N \in \{3, 4, 5, 6, 7, 8\}$
      \item[] $H = 25$
      \item[] $\Delta t = 0.1$
      \item[] $T^{sim} = 2000$
      \item[] $\Delta t^{sim} = 0.01$
      \item[] $H^{ghost} = 20$
      \item[] $Q_\mathbf{\bar{u}} = \left( \begin{smallmatrix} 0.005 & 0 & 0 \\ 0 & 0.005 & 0 \\ 0 & 0 & 100000 \end{smallmatrix} \right) $
      \item[] $r^{rob} = 0.35$
      \item[] $r^{ped} = 0.3$
      \item[] $d^{safe} = 0.3$ 
      \item[] $P^{col} = 0.01$
      \item[] $\varepsilon = 0.1$
      \item[] $\ell^{vis} = 5$ 
      \item[] $\varphi^{vis} = 2\pi$ 
      \item[] $Q_\mathbf{u} = \left( \begin{smallmatrix} 1&0 \\ 0&1 \end{smallmatrix} \right) $
      \item[] $Q_{\mathbf{ED}} = 500$
      \item[] $Q_{\mathbf{MD}} = 1000$,
    \end{itemize}
\end{multicols}
\noindent$Q_{\mathbf{r}} = 100$ if an additional cost component is Euclidean, otherwise $Q_{\mathbf{r}} = 1000$. We make the assumption that the robot is imperceptible to pedestrians, and therefore, they do not respond to its presence. The unicycle kinematics model described in Section \ref{sec_methods} is used to model the robot in both the controller optimization problem and the simulation model. However, there is a difference in the time intervals used- $\Delta t = 0.1$ in the optimization problem and $\Delta t^{sim} = 0.01$ in the simulation model. The controller is invoked every $\Delta t = 0.1$ time interval within the simulation model, which is known as a 'sample and hold' system. To enhance the navigation capabilities, we implemented the \textit{ghost-pedestrian} feature in PyMiniSim. This feature enables the robot to continue tracking the pedestrian using his last trajectory prediction when the pedestrian leaves the robot's field of view, up to $H^{ghost}$ steps.



\subsection{Results and Analysis}
\begin{figure*}
  \includegraphics[width=\textwidth]{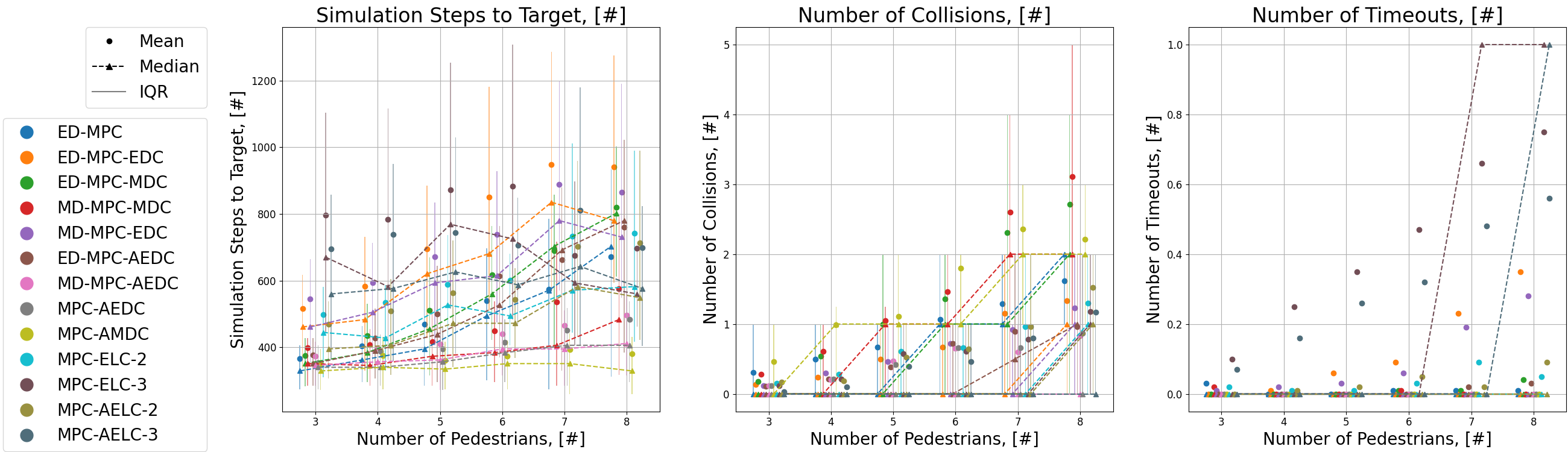}
  \caption{Statistical results for the Circular Crossing Scenario.}
  \label{circular_crossing}
\end{figure*}

\begin{figure*}
  \includegraphics[width=\textwidth]{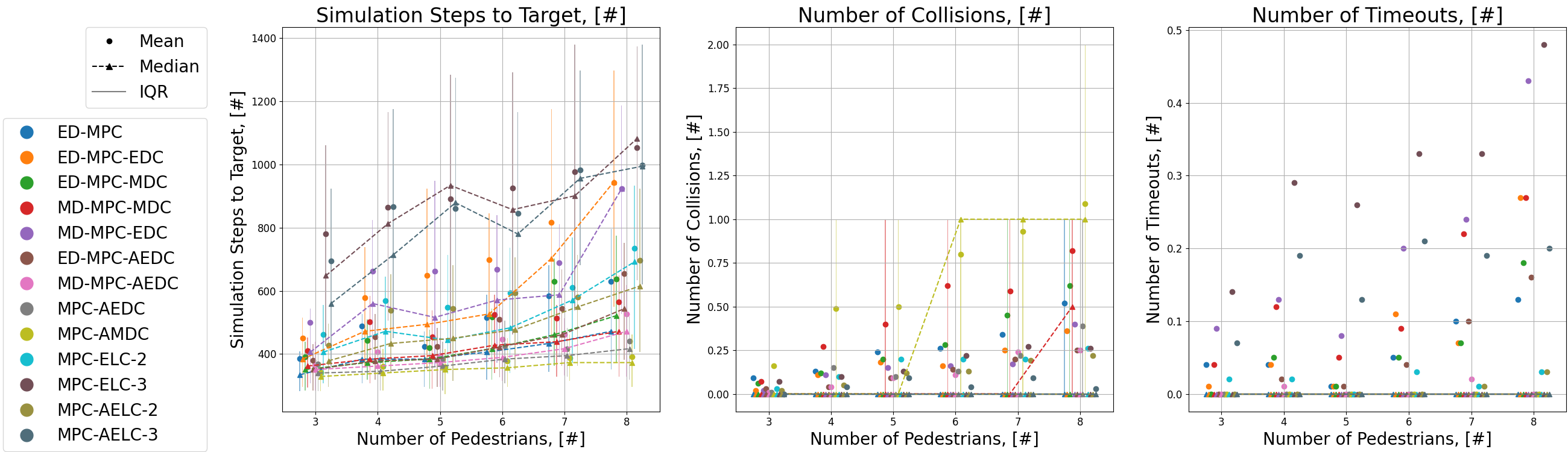}
  \caption{Statistical results for the Random Crossing Scenario.}
  \label{random_crossing}
\end{figure*}

\begin{figure*}
  \includegraphics[width=\textwidth]{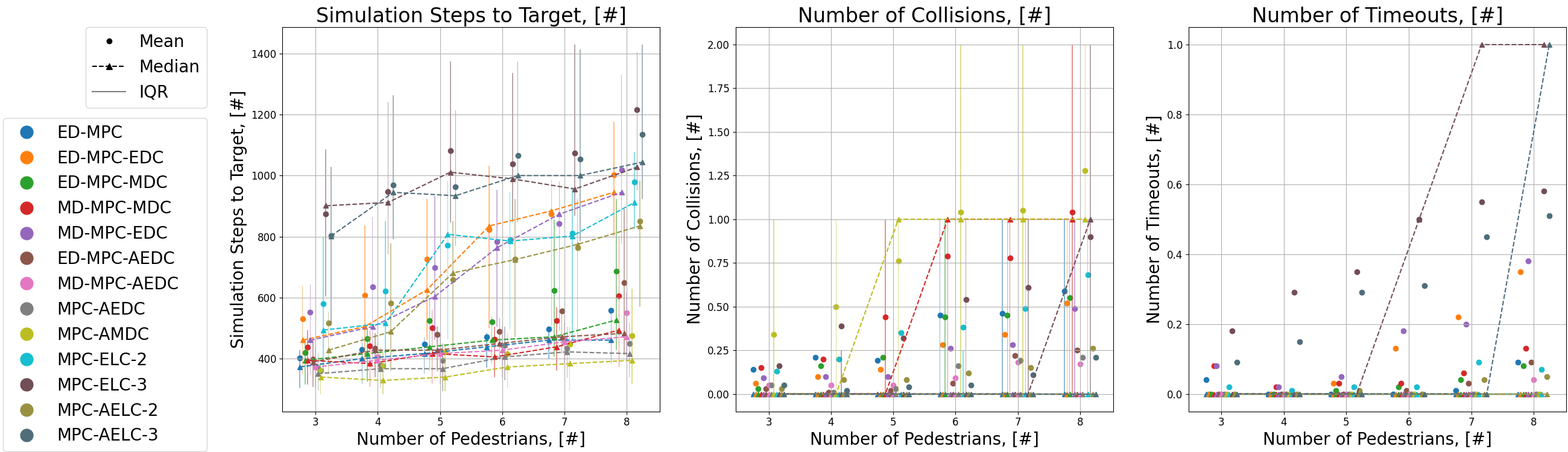}
  \caption{Statistical results for the Parallel Crossing Scenario.}
  \label{parallel_crossing}
\end{figure*}

Results of the experiments are represented by the statistics, collected for each of the three proposed types of the scenarios, showed at Fig. \ref{circular_crossing}, \ref{random_crossing} and \ref{parallel_crossing} with means, medians and interquartile ranges (IQR). Using this data, we provide both scenario-specific analysis and derive general conclusions on the practical applications of the proposed controllers.

According to our observations, the \textit{circular crossing} scenario (Fig. \ref{circular_crossing}) is the most challenging scenario in practice, since when the goal is sampled inside the inner circle, robot needs to reach it as fast as possible until it becomes cramped by the pedestrians; when the goal is sampled outside the inner circle, robot needs to carefully break out of it. 
In terms of the number of collisions, for the lowest number of pedestrians all methods perform similarly. For the larger numbers, we can observe degradation of several uncertainty-aware methods and methods that do not employ adaptive constraints. Generally, for this scenario good performance in terms of collisions is achieved by MD-MPC-AEDC, MPC-AEDC, ED-MPC-AEDC, MPC-ELC-3 and MPC-AELC-3. In terms of numbers of simulation steps and timeouts, we see that ellipsoid constraints based method with largest number of standard deviations (MPC-ELC-3) tend to be much more conservative, and usage of the adaptive constraints (MPC-AELC-3) partially tackles this issue. With smaller number of standard deviations (MPC-ELC-2 and MPC-AELC-3), ellipsoid constraints based methods show level of conservative much closer to the other methods mentioned above. For the hardest case in this scenario, we provide detailed results in Table \ref{circular-results}. For this case, we highlight performance of MD-MPC-AEDC and MPC-AEDC approaches.

The \textit{random crossing} scenario (Fig. \ref{random_crossing}) tends to be a 'medium-complexity' problem for the controllers. We again see the trend of degrading performance of the Mahalanobis and Euclidean non-adaptively constrained controllers. Comparing the hardest cases of 7-8 pedestrians, we can see that good performance is shown by MPC-ELC-2, MPC-ELC-3, MPC-AELC-2, MPC-AELC-3, MD-MPC-AEDC, ED-MPC-AEDC. In terms of simulation steps and timeouts, a gap between MPC-AELC-2 and both MPC-AELC-3 and ED-MPC-AEDC can be seen, same for the MPC-ELC-2 and MPC-ELC-3.

While visually looking like a relatively simple problem, the \textit{parallel crossing} scenario (Fig. \ref{parallel_crossing}) still tends to be a challenging problem for the controllers, especially when the robot becomes close to the two pedestrians approaching each other. The trend of degrading performance of the Mahalanobis and Euclidean non-adaptively constrained controllers can be seen again. Comparing the hardest cases of 6-8 pedestrians, we can see that good performance is shown by MPC-AEDC, MD-MPC-AEDC, MPC-AELC-2, MPC-AELC-3 and ED-MPC-AEDC. Performance in terms of simulation steps and timeouts is similar to the previous case, still we see a huge gap between MPC-AELC-2 and MPC-AELC-3 which gives insight on the influence of the number of standard deviations on the agility.


Based on our findings for each of the scenarios, we can propose following conclusions-
\begin{enumerate}
    \item \textit{Adaptive constraints are the crucial part for MPC-based methods}. We see that leading controllers employ the concept of adaptive constraints. We also observed that adaptive constraints in some cases make controllers more stable.
    \item \textit{Designing uncertainty-aware MPC components is still a hard task}. Poor performance of the methods that use non-adaptive Mahalanobis distance based constraints tells that approximation introduced in \ref{eq:MDC-inequality} is too coarse approximation, and, as was discussed in \ref{sec_related_objectives}, introducing more precise approximations can be a tricky task. On the other hand, chance constraints require tuning to find trade-off between safety and agility. Still, both chance constraints and methods employing Mahalanobis distance based cost showed their potential in making the system safer.
\end{enumerate}

\begin{table}[]
\centering
\caption{Circular Crossing Scenario Results, $N=8$.}
\label{circular-results}
\resizebox{\columnwidth}{!}{%
\begin{tabular}{llll}
\toprule
& \textbf{Simulation Steps to Target} & \textbf{Number of Collisions} & \textbf{Number of Timeouts} \\ 
\textbf{Controller Name} & \multicolumn{3}{c}{ Q1 | Median | Mean | Q2 } \\
\midrule
ED-MPC         & 312.5 | 703.0 | 671.6 | 934.0   &  1.0 | 2.0 | 1.62 | 2.0  & 0.0 | 0.0 | 0.01 | 0.0  \\         
ED-MPC-EDC     & 593.0 | 780.0 | 940.9 | 1275.0  &  0.0 | 1.0 | 1.33 | 2.0  & 0.0 | 0.0 | 0.35 | 1.0  \\      
ED-MPC-MDC     & 612.3 | 802.0 | 819.6 | 1002.8  &  0.0 | 2.0 | 2.71 | 4.0  & 0.0 | 0.0 | 0.04 | 0.0  \\        
MD-MPC-MDC     & 318.0 | 483.0 | 575.8 | 791.0   &  1.0 | 2.0 | 3.11 | 5.0  & 0.0 | 0.0 | 0.00 | 0.0   \\   
MD-MPC-EDC     & 549.0 | 730.5 | 865.5 | 1189.8  &  0.0 | 1.0 | 1.23 | 2.0  & 0.0 | 0.0 | 0.28 | 1.0  \\       
ED-MPC-AEDC    & 384.0 | 780.0 | 760.9 | 1022.0  &  0.0 | 1.0 | 0.96 | 2.0  & 0.0 | 0.0 | 0.03 | 0.0  \\       
\textbf{MD-MPC-AEDC}    & \textbf{315.3} | \textbf{411.5} | \textbf{496.3} | \textbf{562.8}   &  \textbf{0.0} | \textbf{0.0} | \textbf{0.83} | \textbf{2.0}  & \textbf{0.0} | \textbf{0.0} | \textbf{0.00} | \textbf{0.0}   \\
\textbf{MPC-AEDC}       & \textbf{296.0} | \textbf{406.0} | \textbf{483.0} | \textbf{538.0}   &  \textbf{0.0} | \textbf{0.0} | \textbf{0.87} | \textbf{2.0}  & \textbf{0.0} | \textbf{0.0} | \textbf{0.00} | \textbf{0.0}  \\
MPC-AMDC       & 260.3 | 329.0 | 380.5 | 430.8   &  1.0 | 2.0 | 2.21 | 3.0  & 0.0 | 0.0 | 0.00 | 0.0   \\   
MPC-ELC-2      & 417.0 | 582.0 | 741.9 | 989.0   &  0.0 | 1.0 | 1.30 | 2.0  & 0.0 | 0.0 | 0.05 | 0.0  \\    
MPC-ELC-3      & 461.0 | 560.0 | 696.4 | 703.0   &  0.0 | 1.0 | 1.18 | 2.0  & 0.8 | 1.0 | 0.75 | 1.0 \\   
MPC-AELC-2     & 422.5 | 549.0 | 713.2 | 989.0   &  0.0 | 1.0 | 1.52 | 2.0  & 0.0 | 0.0 | 0.09 | 0.0  \\     
MPC-AELC-3     & 403.3 | 576.5 | 699.0 | 824.0   &  0.0 | 0.0 | 1.17 | 2.0  & 0.0 | 1.0 | 0.56 | 1.0  \\    
\bottomrule
\end{tabular}%
}
\end{table}

%% file: conclusion.tex
\section{Conclusion}\label{sec_concl}
In this work, we studied several approaches for designing socially-aware MPC in both uncertainty-unaware and uncertainty-aware settings. We provided their comprehensive evaluation which includes the development of the simulation environment, design of the social scenarios, collection of the statistics on controllers' performance and analysis of those results. We derive several conclusions which may help developers and researchers to decide when embedding uncertainty-awareness may work efficiently and when simpler uncertainty-unaware controllers may provide the same or even better performance. Further directions for our study is to explore uncertainty-unaware and uncertainty-aware MPC implementations with numerical and sampling-based solvers and conducting experiments on the real robotic platform.
